\documentclass{article} 
\usepackage{nips14submit_e,times}
\usepackage{url}

\usepackage{graphicx} 
\usepackage{subfigure}
\usepackage{amsmath,amssymb}
\usepackage{multirow}
\usepackage{wrapfig}

\usepackage[numbers]{natbib}

\usepackage[ruled,vlined,linesnumbered]{algorithm2e}

\newcommand{\cut}[1]{}

\title{Do Deep Nets Really Need to be Deep? \\
*** \\
Draft for NIPS 2014 (not camera ready copy) \\
***}

\author{
Lei Jimmy Ba \\
University of Toronto\\
\texttt{jimmy@psi.utoronto.ca}
\And
Rich Caruana \\
Microsoft Research\\
\texttt{rcaruana@microsoft.com} 
}

\nipsfinalcopy 

\begin{document}

\maketitle

\begin{abstract}

Currently, deep neural networks are the state of the art on problems such as
speech recognition and computer vision.  In this paper we empirically demonstrate that shallow
feed-forward nets can learn the complex functions previously learned by deep
nets and achieve accuracies previously only achievable with deep models.
Moreover, in some cases the shallow neural nets can learn these deep functions
using the same number of parameters as the original deep models.  On the TIMIT
phoneme recognition and CIFAR-10 image recognition tasks, shallow nets can be
trained that perform similarly to complex, well-engineered, deeper
convolutional architectures.

\end{abstract}

\section{Introduction}

You are given a training set with 1M labeled points.  When you train
a shallow neural net with one fully-connected feed-forward hidden layer
on this data you obtain 86\% accuracy on test data.  When you train a
deeper neural net as in \cite{abdel2012applying} consisting 
of a convolutional layer, pooling layer,
and three fully-connected feed-forward layers on the same data you
obtain 91\% accuracy on the same test set.

What is the source of this improvement?  Is the 5\% increase in accuracy of
the deep net over the shallow net because: 
a) the deep net has more parameters; 
b) the deep net can learn more complex functions given the same number of parameters; 
c) the deep net has better bias and learns more interesting/useful functions (e.g., because the deep net is {\em deeper} it learns hierarchical representations \cite{dauphin2013big}); 
d) nets without convolution can't easily learn what nets with convolution can learn; 
e) current learning algorithms and regularization methods work better with deep architectures than shallow architectures\cite{erhan2010does}; 
f) all or some of the above; 
g) none of the above?



There have been attempts to answer the question above. It has been shown that
deep nets coupled with unsupervised layer-by-layer pre-training
technique\cite{hinton2006reducing} \cite{vincent2010stacked} work well.  In
\cite{erhan2010does}, the authors show that depth combined with pre-training
provides a good prior for model weights, thus improving generalization. There
is well-known early theoretical work on the representational capacity of neural
nets. For example, it was proved that a network with a large enough single
hidden layer of sigmoid units can approximate any decision
boundary\cite{cybenko1989approximation}. Empirical work, however, shows that it
is difficult to train shallow nets to be as accurate as deep nets. For vision
tasks, a recent study on deep convolutional nets suggests that deeper models
are preferred under a parameter budget \cite{eigen2013understanding}. In
\cite{dauphin2013big}, the authors trained shallow nets on SIFT features to
classify a large-scale ImageNet dataset and showed that it is challenging to
train large shallow nets to learn complex functions. And in
\cite{seide2011conversational}, the authors show that deeper models are more
competitive than shallow models in speech acoustic modeling. 

In this paper we provide empirical evidence that shallow nets are capable of
learning the same function as deep nets, and in some cases with the same number
of parameters as the deep nets.  We do this by first training a
state-of-the-art deep model, and then training a shallow model to mimic the
deep model.  The mimic model is trained using the model compression scheme
described in the next section.  Remarkably, with model compression we are able
to train shallow nets to be as accurate as some deep models, even though we are
not able to train these shallow nets to be as accurate as the deep nets when
the shallow nets are trained directly on the original labeled training data.
If a shallow net {\em with the same number of parameters} as a deep net can
learn to mimic a deep net with high fidelity, then it is clear that the
function learned by that deep net does not really have to be deep.  


\section{Training Shallow Nets to Mimic Deep Nets}

\subsection{Model Compression}

The main idea behind model compression is to train a compact model to
approximate the function learned by a larger, more complex model.  For example,
in \cite{buciluǎ2006model}, a single neural net of modest size could be trained
to mimic a much larger ensemble of models --- although the {\em small} neural
nets contained 1000 times fewer parameters, often they were just as accurate as
the ensembles they were trained to mimic.  Model compression works by passing
unlabeled data through the large, accurate model to collect the scores
produced by that model.  This synthetically labeled data is then used to train
the smaller mimic model.  The mimic model is not trained on the original
labels---it is trained to learn the {\em function} that was learned by the
larger model.  If the compressed model learns to mimic the large model
perfectly it makes exactly the same predictions and mistakes as the complex
model.

Surprisingly, often it is not (yet) possible to train a small neural net on the
original training data to be as accurate as the complex model, {\em nor as
accurate as the mimic model}.  Compression demonstrates that a small neural net
could, {\em in principle}, learn the more accurate function, but current
learning algorithms are unable to train a model with that accuracy from the
original training data; instead, we must train the complex intermediate model
first and then train the neural net to mimic it.  Clearly, when it is possible
to mimic the function learned by a complex model with a small net, the function
learned by the complex model wasn't truly too complex to be learned by a small
net.  This suggests to us that the complexity of a learned model, and the size
of the representation best used to learn that model, are different things.  In
this paper we apply  model compression to train shallow neural nets to mimic
deeper neural nets, thereby demonstrating that deep neural nets may not need to
be deep.

\subsection{Mimic Learning via Regressing Logit with L2 Loss}
\label{sec:regressionL2Loss}

On both TIMIT and CIFAR-10 we train shallow mimic nets using data labeled by 
either a deep net, or an ensemble of deep nets, trained on the original TIMIT
or CIFAR-10 training data. The deep models are trained in the usual way using
softmax output and cross-entropy cost function. The shallow mimic models,
however, instead of being trained with cross-entropy on the 183 $p$ values
where $p_k = {e^{z_k} / \sum_j e^{z_j}}$ output by the softmax layer from the
deep model, are trained directly on the 183 log probability values $z$, also
called logit, {\em before} the softmax activation.

Training on these logit values makes learning easier for the shallow net by placing emphasis on all prediction targets. Because the logits capture the logarithm relationships between the probability predictions, a student model trained on logits has to learn all of the additional fine detailed relationships between labels that is not obvious in the probability space yet was learned by the teacher model. For example, assume there are three targets that the teacher predicts with probability $[2e-9, 4e-5, 0.9999]$. If we use these probabilities as prediction targets directly to minimize a cross entropy loss function, the student will focus on the third target and easily ignore the first and second target. Alternatively, one can extract the logit prediction from the teacher model and obtain our new targets $[10, 20, 30]$. The student will learn to regress the third target, yet it still learns the first and second target along with their relative difference. The logit values provide richer information to student to mimic the exact behaviours of a teach model.  Moreover, consider a second training case where the teacher predicts logits $[-10, 0, 10]$.  After softmax, these logits yield the same predicted probabilities as $[10, 20, 30]$, yet clearly the teacher has learned internally to model these two cases very differently.  By training the student model on the logits directly, the student is better able to learn the internal model learned by the teacher, without suffering from the information loss that occurs after passing through the logits to probability space.


We formulate the SNN-MIMIC learning objective function as a regression problem given 
training data \{($x^{(1)}, z^{(1)}$),...,($x^{(T)}, z^{(T)}$) \}:

\begin{gather}
\mathcal{L}(W, \beta) = {1 \over 2T}\sum_{t} ||g(x^{(t)};W, \beta) - z^{(t)}||^2_2, 
\label{eq:cost}
\end{gather}

where, $W$ is the weight matrix between input features $x$ and hidden layer,
$\beta$ is the weights from hidden to output units, $g(x^{(t)};W, \beta)  =
\beta f(Wx^{(t)})$ is the model prediction on the $t^{th}$ training data point
and $f(\cdot)$ is the non-linear activation of the hidden units. The parameters
$W$ and $\beta$ are updated using standard error back-propagation algorithm and
stochastic gradient descent with momentum. 

We have also experimented with other different mimic loss function, such as
minimizing the KL divergence $\text{KL}(p_{\text{teacher}}\|p_{\text{student}})$ cost
function and L2 loss on the probability. Logits regression outperforms all the
other loss functions and is one of the key technique for obtaining the results
in the rest of this paper. We found that normalizing the logits from the
teacher model, by subtracting the mean and dividing the standard deviation of
each target across the training set, can improve the L2 loss slightly during
training. Normalization is not crucial for obtaining a good student model.

\subsection{Speeding-up Mimic Learning by Introducing a Linear Layer}
\label{sec:linearLayerSpeedup}

To match the number of parameters in a deep net, a shallow net has to have more
non-linear hidden units in a single layer to produce a large weight matrix $W$.
When training a large shallow neural network with many hidden units, we find it
is very slow to learn the large number of parameters in the weight matrix
between input and hidden layers of size $O(HD)$, where $D$ is input feature
dimension and $H$ is the number of hidden units. Because there are many highly
correlated parameters in this large weight matrix gradient
descent converges slowly. We also notice that during learning, shallow nets
spend most of the computation in the costly matrix multiplication of the input
data vectors and large weight matrix. The shallow nets eventually learn
accurate mimic functions, but training to convergence is very slow (multiple
weeks) even with a GPU.


We found that introducing a bottleneck linear layer with $k$ {\em linear} hidden units
between the input and the non-linear hidden layer sped up learning
dramatically: we can factorize the weight matrix $W\in\mathbb{R}^{H \times D}$
into the product of two low rank matrices, $U\in\mathbb{R}^{H \times k}$ and
$V\in\mathbb{R}^{k \times D}$, where $k<<D,H$. The new cost function can be
written as:
\begin{gather}
\mathcal{L}(U, V, \beta) = {1 \over 2T}\sum_{t} ||\beta f(UVx^{(t)}) - z^{(t)}||^2_2 
\label{eq:cost_linear}
\end{gather}
The weights $U$ and $V$ can be learnt by back-propagating through the linear
layer. This re-parameterization of weight matrix $W$ not only increases the
convergence rate of the shallow mimic nets, but also reduces
memory space from $O(HD)$ to $O(k(H+D))$. 

Factorizing weight matrices has been previously explored in
\cite{sainath2013low} and \cite{xue2013restructuring}. While these prior works
focus on using matrix factorization in the last output layer, our method is
applied between input and hidden layer to improve the convergence speed during
training.

The reduced memory usage enables us to train large shallow models that were
previously infeasible due to excessive memory usage. \cut{Because the linear layer
can always be absorbed into the weight matrix $W$, even after adding the linear
layer the model has similar representational power as the original
shallow net before adding the linear layer.} The linear bottle neck can only
reduce the representational power of the network, and it can always be absorbed
into a signle weight matrix $W$.

\section{TIMIT Phoneme Recognition}

The TIMIT speech corpus has 462 speakers in the training set. There is a
separate development set for cross-validation including 50 speakers, and a
final test set with 24 speakers. The raw waveform audio data were pre-processed
using 25ms Hamming window shifting by 10ms to extract Fourier-transform-based
filter-banks with 40 coefficients (plus energy) distributed on a mel-scale,
together with their first and second temporal derivatives. We included +/- 7
nearby frames to formulate the final 1845 dimension input vector. The data
input features were normalized by subtracting the mean and dividing by the
standard deviation on each dimension. All 61 phoneme labels are represented in
tri-state, i.e., 3 states for each of the 61 phonemes, yielding target label
vectors with 183 dimensions for training.  At decoding time these are mapped to
39 classes as in \cite{lee1989speaker} for scoring. 

\subsection{Deep Learning on TIMIT}


Deep learning was first successfully applied to speech recognition
in \cite{mohamed2012acoustic}. 
We follow the same framework and train two deep models on TIMIT, 
DNN and CNN. DNN is a deep neural net consisting of three 
fully-connected feedforward hidden layers consisting of 2000 
rectified linear units (ReLU) \cite{nair2010rectified} per layer. 
CNN is a deep neural net consisting of a convolutional layer and 
max-pooling layer followed by three hidden layers containing 2000 
ReLU units \cite{abdelexploring}. The CNN was trained using the same convolutional
architecture as in \cite{deng2013recent}.  We also formed an ensemble of nine 
CNN models, ECNN. 

The accuracy of DNN, CNN, and ECNN on the final test set are shown in
Table~\ref{tb:comparison}.  The error rate of the convolutional deep net (CNN) is about
2.1\% better than the deep net (DNN).  The table also shows the accuracy of
shallow neural nets with 8000, 50,000, and 400,000 hidden units (SNN-8k, SNN-50k, and SNN-400k) 
trained on the original training data.  Despite having up to 10X as many
parameters as DNN, CNN and ECNN, the shallow models are 1.4\% to 2\% less
accurate than the DNN, 3.5\% to 4.1\% less accurate than the CNN, and 4.5\% to 5.1\% less accurate than the ECNN.

\subsection{Learning to Mimic an Ensemble of Deep Convolutional TIMIT Models}

The most accurate {\em single} model we trained on TIMIT is the deep convolutional architecture in \cite{deng2013recent}.   Because we have no unlabeled data from the TIMIT distribution, we are forced to use the same 1.1M points in the train set as unlabeled data for compression by throwing away their labels.\footnote{That SNNs can be trained to be as accurate as DNNs using {\em only} the original training data data highlights that it {\em should} be possible to train accurate SNNs on the original train data given better learning algorithms.} Re-using the train set reduces the accuracy of the mimic models, increasing the gap between the teacher and mimic models on test data: model compression works best when the unlabeled set is much larger than the train set, and when the unlabeled samples do not fall on train points where the teacher model is more likely to have overfit. To reduce the impact of the gap caused by performing compression with the original train set, we train the student model to mimic a more accurate ensemble of deep convolutional models.

We are able to train a more accurate model on TIMIT by forming an ensemble of 9 deep, convolutional neural nets, each trained with somewhat different train sets, and with architectures with different kernel sizes in the convolutional layers.   We used this very accurate model, ECNN, as the teacher model to label the data used to train the shallow mimic nets.  As described in Section~\ref{sec:regressionL2Loss}, the logits (log probability of the predicted values) from each CNN in the ECNN model are averaged and the average logits are used as final regression targets to train the mimic SNNs. 

We trained shallow mimic nets with 8k (SNN-MIMIC-8k) and 400k (SNN-MIMIC-400k) hidden units on the re-labeled 1.1M training points.  As described in Section~\ref{sec:linearLayerSpeedup}, both mimic models have 250 linear units between the input and non-linear hidden layer to speed up learning --- preliminary experiments suggest that for TIMIT there is little benefit from using more than 250 linear units.

\subsection{Compression Results For TIMIT}

The bottom of Table~\ref{tb:comparison} shows the accuracy of shallow mimic nets with 8000 ReLUs 
and 400,000 ReLUs (SNN-MIMIC-8k and -400k) trained with model compression to mimic 
the ECNN.
Surprisingly, shallow nets are able to perform as well as their deep
counter-parts when trained with model compression to mimic a more accurate
model. A neural net with one hidden layer (SNN-MIMIC-8k) can be trained to
perform as well as a DNN with a similar number of parameters.
Furthermore, if we increase the number of hidden units in the shallow net from
8k to 400k (the largest we could train), we see that a neural net with one
hidden layer (SNN-MIMIC-400k) can be trained to perform comparably to a CNN
even though the SNN-MIMIC-400k net has no convolutional or pooling layers.
This is interesting because it suggests that a large single hidden<
layer without a topology custom designed for the problem is able to reach the
performance of a deep convolutional neural net that was carefully engineered
with prior structure and weight sharing {\em without any increase in the number
of training examples}, even though the same architecture trained on the original data could not.

\begin{table*}[t]%
\centering
\begin{tabular}{|l|c|c|c|c|c|} \hline
 & \multirow{2}{*}{Architecture} & \multirow{2}{*}{\# Param.} & \multirow{2}{*}{\# Hidden units} & \multirow{2}{*}{PER} \\
 & & & &\\ \hline \hline
\multirow{2}{*}{SNN-8k} & 8k + dropout & \multirow{2}{*}{$\sim$12M}& \multirow{2}{*}{$\sim$8k} & \multirow{2}{*}{23.1\%}
    \\ & trained on original data & & &\\ \hline
\multirow{2}{*}{SNN-50k} & 50k + dropout & \multirow{2}{*}{$\sim$100M}& \multirow{2}{*}{$\sim$50k} & \multirow{2}{*}{23.0\%}
    \\ & trained  on original data & & &\\ \hline
\multirow{2}{*}{SNN-400k} & 250L-400k + dropout & \multirow{2}{*}{$\sim$180M}& \multirow{2}{*}{$\sim$400k} & \multirow{2}{*}{23.6\%}
    \\ & trained on original data & & &\\ \hline
\multirow{2}{*}{DNN} & 2k-2k-2k + dropout  & \multirow{2}{*}{$\sim$12M}& \multirow{2}{*}{$\sim$6k} & \multirow{2}{*}{21.9\%}
    \\  & trained on original data & & &\\ \hline
\multirow{2}{*}{CNN} & c-p-2k-2k-2k + dropout & \multirow{2}{*}{$\sim$13M}& \multirow{2}{*}{$\sim$10k} & \multirow{2}{*}{\bf{19.5\%}}
    \\  & trained on original data & & &\\ \hline 
\multirow{2}{*}{ECNN} & \multirow{2}{*}{ensemble of 9 CNNs} & \multirow{2}{*}{$\sim${125M}}& \multirow{2}{*}{$\sim$90k} & \multirow{2}{*}{{\bf{18.5\%}}} \\ 
    & & & & \\ \hline \hline
\multirow{2}{*}{SNN-MIMIC-8k} & 250L-8k & \multirow{2}{*}{$\sim$12M}& \multirow{2}{*}{$\sim$8k} & \multirow{2}{*}{\bf{21.6\%}}
    \\ & no convolution or pooling layers & & & \\ \hline
\multirow{2}{*}{SNN-MIMIC-400k} & 250L-400k & \multirow{2}{*}{$\sim$180M}& \multirow{2}{*}{$\sim$400k} & \multirow{2}{*}{\bf{20.0\%}}
    \\ & no convolution or pooling layers & & & \\ \hline
\end{tabular}
\caption{Comparison of shallow and deep models: phone error rate (PER) on TIMIT core
test set.}
\label{tb:comparison}
\end{table*}

\begin{figure*}
\hbox{
\includegraphics[width=0.5\textwidth,height=0.35\textwidth]{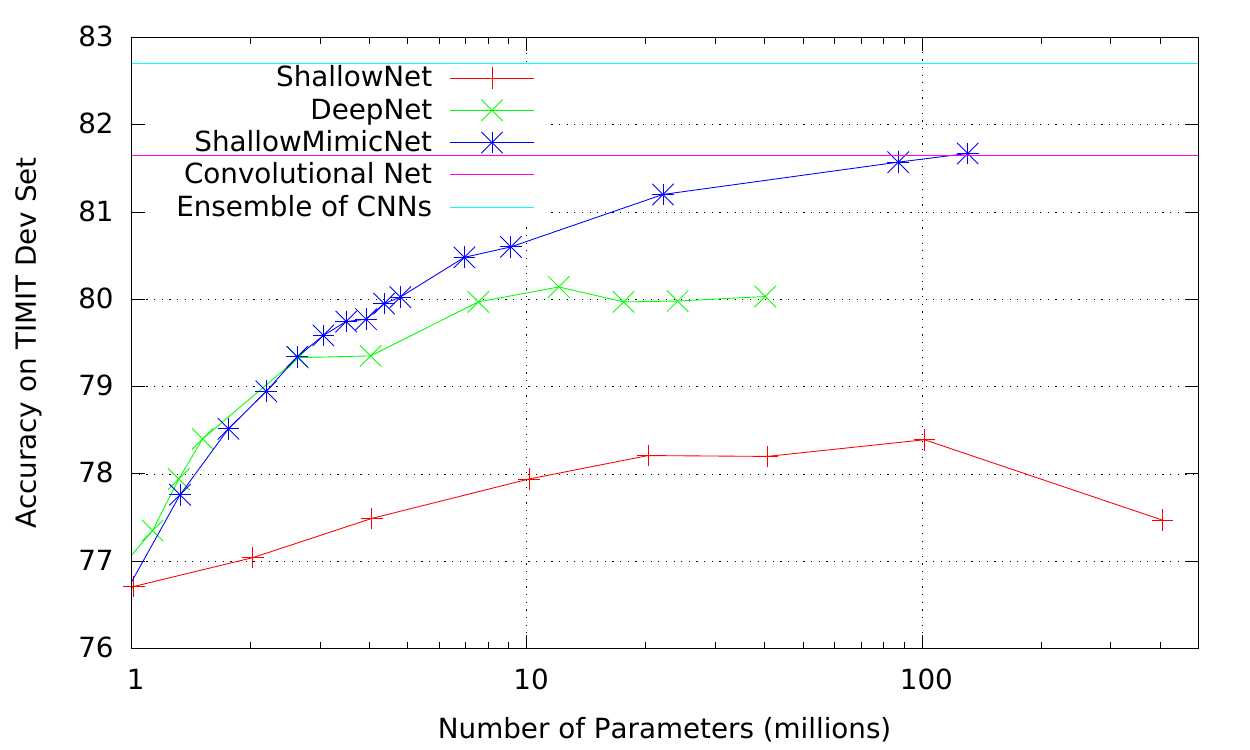}
\includegraphics[width=0.5\textwidth,height=0.35\textwidth]{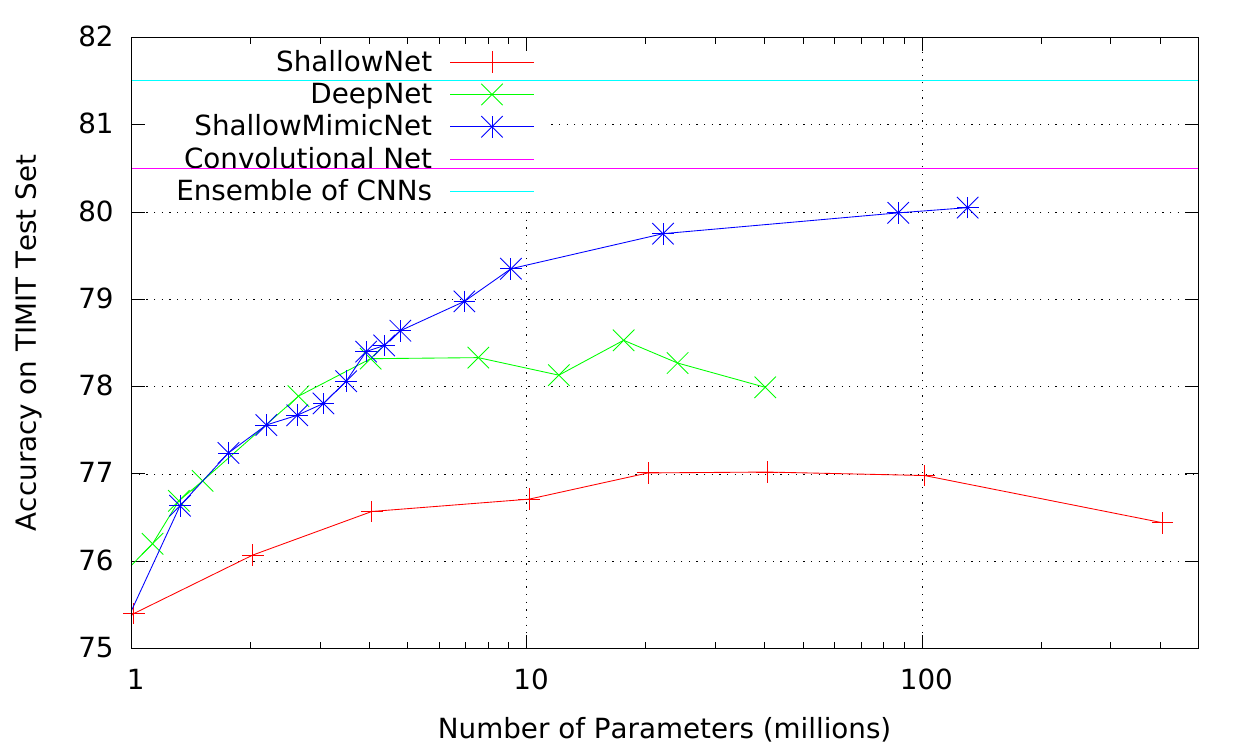}
}
\vspace*{-0.1in}
\caption{Accuracy of SNNs, DNNs, and Mimic SNNs vs. \# of parameters on TIMIT Dev (left) and Test (right) sets.  Accuracy of the CNN and target ECNN are shown as horizontal lines for reference.}
\label{fig:TIMITaccuracyVsParameters}
\end{figure*}

Figure~\ref{fig:TIMITaccuracyVsParameters} shows the accuracy of shallow nets 
and deep nets trained on the original TIMIT 1.1M data, and shallow mimic nets 
trained on the ECNN targets, as a function of the number of parameters in the 
models.  The accuracy of the CNN and the teacher ECNN are shown as horizontal 
lines at the top of the figures.  When the number of parameters is small (about 1 
million), the SNN, DNN, and SNN-MIMIC models all have similar accuracy.  As the 
size of the hidden layers increases and the number of parameters increases, the 
accuracy of a shallow model trained on the original data begins to lag behind.  The 
accuracy of the shallow mimic model, however, matches the accuracy of the DNN until 
about 4 million parameters, when the DNN begins to fall behind the mimic.  The DNN 
asymptotes at around 10M parameters, while the shallow mimic continues to increase 
in accuracy.  Eventually the mimic asymptotes at around 100M parameters to an accuracy 
comparable to that of the CNN.  The shallow mimic never achieves the accuracy of the 
ECNN it is trying to mimic (because there is not enough unlabeled data), but it is able to 
match or exceed the accuracy of deep nets (DNNs) {\em having the same number of 
parameters} trained on the original data.

\section{Object Recognition: CIFAR-10}

To verify that the results on TIMIT generalize to other learning problems and
task domains, we ran similar experiments on the CIFAR-10 Object Recognition
Task\cite{krizhevsky2009learning}. CIFAR-10 consists of a set of natural images
from 10 different object classes: airplane, automobile, bird, cat, deer, dog,
frog, horse, ship, truck. The dataset is a labeled subset of the 80 million tiny
images dataset\cite{torralba200880} and is divided into 50,000 train and 10,000
test images. Each image is 32x32 pixels in 3 color channels, yielding input
vectors with 3072 dimensions. We prepared the data by subtracting the mean and
dividing the standard deviation of each image vector to perform global contrast
normalization. We then applied ZCA whitening to the normalized images. This
pre-processing is the same used in \cite{goodfellow2013maxout}. 

\subsection{Learning to Mimic a Deep {\em Convolutional} Neural Network}

Deep learning currently achieves state-of-the-art accuracies 
on many computer vision problems. The key to this success is deep
convolutional nets with many alternating layers of convolutional, pooling and non-linear units.
Recent advances such as dropout are also important to prevent over-fitting in these deep nets. 

We follow the same approach as with TIMIT:  An ensemble of deep CNN models is
used to label CIFAR-10 images for model compression. The logit predictions from
this teacher model are used as regression targets to train a mimic shallow neural net (SNN).
CIFAR-10 images have a higher dimension than TIMIT (3072 vs. 1845), but the
size of the CIFAR-10 training set is only 50,000 compared to 1.1 million
examples for TIMIT.  Fortunately, unlike TIMIT, in CIFAR-10 we have access to
unlabeled data from a similar distribution by using the super set of CIFAR-10:
the 80 million tiny images dataset.  We add the first 1 million images from the
80 million set to the original 50,000 CIFAR-10 training images to create a
1.05M mimic training (transfer) set.


CIFAR-10 images are raw pixels for objects viewed from many different angles
and positions, whereas TIMIT features are human-designed filter-bank features.
In preliminary experiments we observed that non-convolutional nets do not
perform well on CIFAR-10 no matter what their depth. Instead of raw pixels, the
authors in \cite{dauphin2013big} trained their shallow models on the SIFT
features. Similarly, \cite{eigen2013understanding} used a base
convolution and pooling layer to study different deep architectures.  We follow the approach
in \cite{eigen2013understanding} to allow our shallow models
to benefit from convolution while keeping the models as shallow as possible,
and introduce a single layer of convolution and pooling in our shallow mimic
models to act as a feature extractor to create invariance to small translations
in the pixel domain. The SNN-MIMIC models for CIFAR-10 thus consist of a
convolution and max pooling layer followed by fully connected 1200 linear units
and 30k non-linear units. As before, the linear units are there only to speed
learning; they do not increase the model's representational power and can be
absorbed into the weights in the non-linear layer after learning.

\begin{table*}[t]%
\centering
\begin{tabular}{|l|c|c|c|c|c|} \hline
 & \multirow{2}{*}{Architecture} & \multirow{2}{*}{\# Param.} & \multirow{2}{*}{\# Hidden units} & \multirow{2}{*}{Err.} \\
 & & & &\\ \hline \hline
    \multirow{2}{*}{DNN} & \multirow{2}{*}{2000-2000 + dropout} & \multirow{2}{*}{$\sim$10M} & \multirow{2}{*}{4k} & \multirow{2}{*}{57.8\%}
    \\ & & & & \\ \hline
    \multirow{2}{*}{SNN-30k} & 128c-p-1200L-30k & \multirow{2}{*}{$\sim$70M} & \multirow{2}{*}{$\sim$190k} & \multirow{2}{*}{21.8\%}
    \\ & + dropout input\&hidden & & & \\ \hline
    single-layer & 4000c-p & \multirow{2}{*}{$\sim$125M} & \multirow{2}{*}{$\sim$3.7B} & \multirow{2}{*}{18.4\%}
    \\ feature extraction  & followed by SVM & & & \\ \hline
CNN\cite{hinton2012improving}  & 64c-p-64c-p-64c-p-16lc  & \multirow{2}{*}{$\sim$10k}& \multirow{2}{*}{$\sim$110k} & \multirow{2}{*}{{15.6\%}}
    \\ (no augmentation) & + dropout on lc & & & \\ \hline 
CNN\cite{zeiler2013stochastic}  & {64c-p-64c-p-128c-p-fc}  & \multirow{3}{*}{$\sim$56k}& \multirow{3}{*}{$\sim$120k} & \multirow{3}{*}{{15.13\%}}
    \\ (no augmentation) & + dropout on fc  & & &
    \\  & and stochastic pooling & & & \\ \hline 
teacher CNN & {128c-p-128c-p-128c-p-1000fc}  & \multirow{3}{*}{$\sim$35k}& \multirow{3}{*}{$\sim$210k} & \multirow{3}{*}{\bf{12.0\%}}
    \\ (no augmentation) & + dropout on fc  & & &
    \\  & and stochastic pooling & & & \\ \hline
ECNN & \multirow{2}{*}{ensemble of 4 CNNs} & \multirow{2}{*}{$\sim$140k}& \multirow{2}{*}{$\sim$840k} & {\bf{\multirow{2}{*}{11.0\%}}} 
    \\ (no augmentation)  & & & & \\ \hline\hline
                  SNN-CNN-MIMIC-30k & 64c-p-1200L-30k  & \multirow{2}{*}{$\sim$54M} & \multirow{2}{*}{$\sim$110k} & \multirow{2}{*}{\bf{15.4\%}}
    \\ trained on a single CNN & with no regularization & & & \\ \hline
                  SNN-CNN-MIMIC-30k & 128c-p-1200L-30k  & \multirow{2}{*}{$\sim$70M} & \multirow{2}{*}{$\sim$190k} & \multirow{2}{*}{\bf{15.1\%}}
    \\ trained on a single CNN & with no regularization & & & \\ \hline
                  SNN-ECNN-MIMIC-30k & 128c-p-1200L-30k & \multirow{2}{*}{$\sim$70M} & \multirow{2}{*}{$\sim$190k} & \multirow{2}{*}{\bf{14.2\%}}
    \\ trained on ensemble& with no regularization & & & \\\hline
\end{tabular}
\caption{Comparison of shallow and deep models: classification error rate on CIFAR-10. 
         Key: c, convolution layer; p, pooling layer; lc, locally connected layer; fc, fully connected layer}
\label{tb:comparison_cifar}
\end{table*}

Results on CIFAR-10 are consistent with those from TIMIT.
Table~\ref{tb:comparison_cifar} shows results for the shallow mimic models, and
for much-deeper convolutional nets. The shallow mimic net trained to mimic the
teacher CNN (SNN-CNN-MIMIC-30k) achieves accuracy comparable to CNNs with
multiple convolutional and pooling layers.  And by training the shallow model
to mimic the ensemble of CNNs (SNN-ECNN-MIMIC-30k), accuracy is improved an
additional 0.9\%.  The mimic models are able to achieve accuracies previously
unseen on CIFAR-10 with models with so few layers. Although the deep
convolution nets have more hidden units than the shallow mimic models, because
of weight sharing, the deeper nets with multiple convolution layers have fewer
parameters than the shallow fully-connected mimic models.  Still, it is
surprising to see how accurate the shallow mimic models are, and that their
performance continues to improve as the performance of the teacher model
improves (see further discussion of this in Section~\ref{sec:followTheLeader}).

\section{Discussion}

\subsection{Why Mimic Models Can Be More Accurate than Training on Original Labels}

It may be surprising that models trained on the prediction targets taken from
other models can be more accurate than models trained on the original labels.
There are a variety of reasons why this can happen:
\begin{itemize}
\item if some labels have errors, the teacher model may eliminate some of these
    errors (i.e., censor the data), thus making learning easier for the
    student: on TIMIT, there are mislabeled frames introduced by the HMM
    forced-alignment procedure.
\item if there are regions in the $p(y|X)$ that are difficult to learn given
    the features, sample density, and function complexity, the teacher may
    provide simpler, soft labels to the student. The complexity in the data 
    set has been {\em washed away} by {\em filtering} the targets through the teacher model.
\item learning from the original hard 0/1 labels can be more difficult than
    learning from the teacher's conditional probabilities: on TIMIT only one of
    183 outputs is non-zero on each training case, but the mimic model sees
    non-zero targets for most outputs on most training cases.  Moreover, the
    teacher model can spread the uncertainty over multiple outputs when it is
    not confident of its prediction. Yet, the teacher model can concentrate the
    probability mass on one (or few) outputs on easy cases. The uncertainty from the teacher
    model is far more informative to guiding the student model than the
    original 0/1 labels. This benefit appears to be further enhanced by
    training on logits.
\end{itemize}

\begin{wrapfigure}{r}{0.5\textwidth}
\hspace{20pt}
\vspace{-45pt}
\includegraphics[width=0.38\textwidth,height=0.35\textwidth]{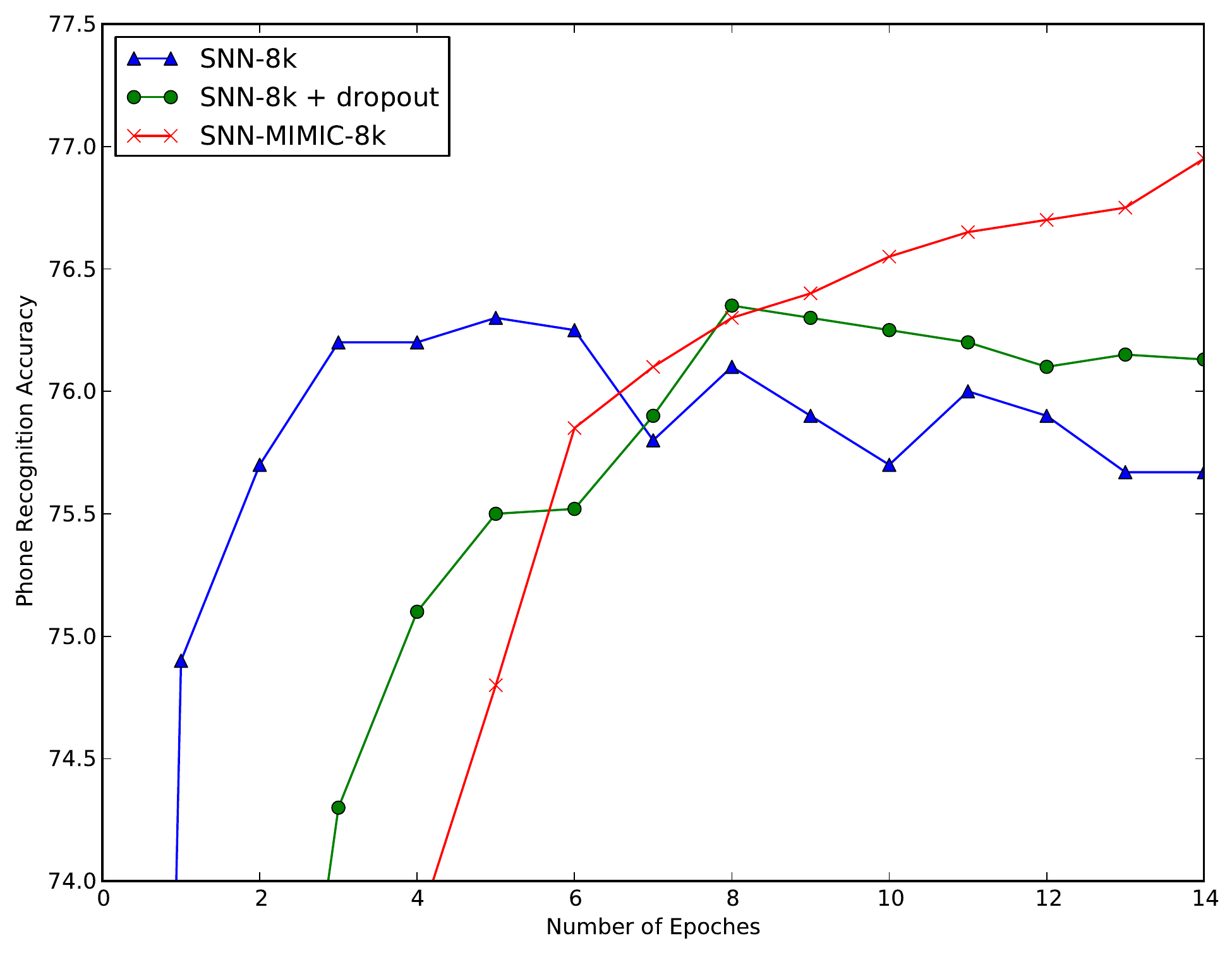}
\label{fig:overfitting}
\vspace{30pt}
\caption{Training  shallow mimic model prevents overfitting. } 
\vspace{-30pt} \end{wrapfigure}

The mechanisms above can be seen as forms of regularization that help prevent
overfitting in the student model.  Shallow models trained on the original
targets are more prone to overfitting than deep models---they begin to overfit
before learning the accurate functions learned by deeper models even with
dropout (see Figure~2\cut{\ref{fig:overfitting}}). If we had more effective
regularization methods for shallow models, some of the performance gap between
shallow and deep models might disappear. Model compression appears to be a form
of regularization that is effective at reducing this gap.

\subsection{The Capacity and Representational Power of Shallow Models}
\label{sec:followTheLeader}

\begin{wrapfigure}{r}{0.5\textwidth}
\hspace{-35pt}
\vspace{-5pt}
\includegraphics[width=0.65\textwidth,height=0.35\textwidth]{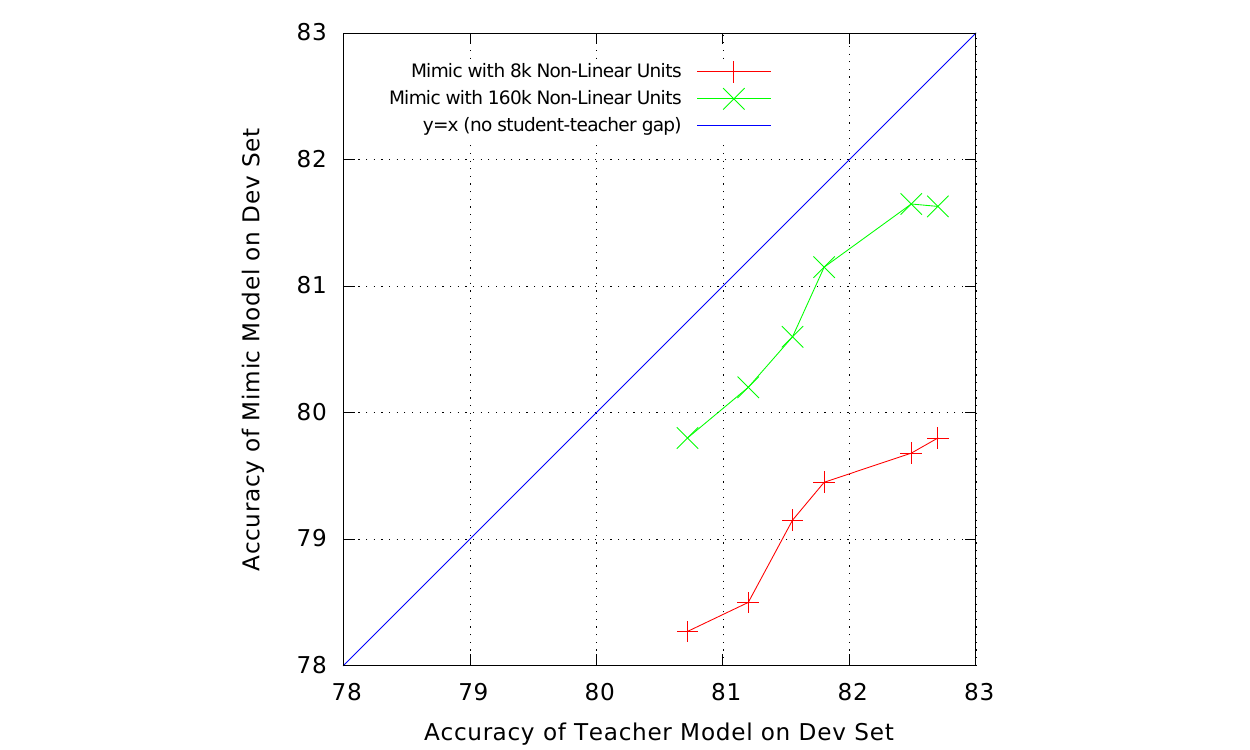}
\label{fig:followTheLeader}
\vspace{-5pt}
\caption{Accuracy of student models continues to improve as accuracy of teacher models improves.} 
\vspace{-10pt} \end{wrapfigure}

Figure~3\cut{\ref{fig:followTheLeader}} shows results of an experiment with TIMIT
where we trained shallow mimic models of two sizes (SNN-MIMIC-8k and
SNN-MIMIC-160k) on teacher models of different accuracies. The two shallow
mimic models are trained on the same number of data points. The only difference
between them is the size of the hidden layer. The x-axis shows the accuracy of
the teacher model, and the y-axis is the accuracy of the mimic models.  Lines
parallel to the diagonal suggest that increases in the accuracy of the teacher
models yield similar increases in the accuracy of the mimic models.  Although
the data does not fall perfectly on a diagonal, there is strong evidence that
the accuracy of the mimic models continues to increase as the accuracy of the
teacher model improves, suggesting that the mimic models are not (yet) running
out of capacity. When training on the same targets, SNN-MIMIC-8k always perform
worse than SNN-MIMIC-160K that has 10 times more parameters. Although there is
a consistent performance gap between the two models due to the difference in size, the
smaller shallow model was eventually able to achieve a performance comparable to the larger
shallow net by learning from a better teacher, and the accuracy of both models continues to increase 
as teacher accuracy increases. This suggests that  
shallow models with a number of parameters comparable to deep models are likely
capable of learning even more accurate functions if a more accurate teacher
and/or more unlabeled data became available.  Similarly, on CIFAR-10 we saw
that increasing the accuracy of the teacher model by forming an ensemble of
deep CNNs yielded commensurate increase in the accuracy of the student model.
We see little evidence that shallow models have limited capacity or
representational power. Instead, the main limitation appears to be the learning
and regularization procedures used to train the shallow models.

\subsection{Parallel Distributed Processing vs. Deep Sequential Processing}
\label{sec:parallel}

Our results show that shallow nets can be competitive with deep models on
speech and vision tasks.  One potential benefit of shallow nets is that
training them scales well with the modern parallel hardware. In our experiments the
deep models usually required 8--12 hours to train on Nvidia
GTX 580 GPUs to reach the state-of-the-art performance on TIMIT and CIFAR-10
datasets. Although some of the shallow mimic models have more parameters than
the deep models, the shallow models train much faster and reach similar
accuracies in only 1--2 hours.

Also, given parallel computational resources, at run-time shallow models can
finish computation in 2 or 3 cycles for a given input, whereas a deep
architecture has to make sequential inference through each of its layers,
expending a number of cycles proportional to the depth of the model. This benefit can be
important in on-line inference settings where data parallelization is not as
easy to achieve as it is in the batch inference setting.  For real-time applications such as
surveillance or real-time speech translation, a model that responds in fewer cycles can be beneficial.

\vspace{-6pt}

\section{Future Work}

The tiny images dataset contains 80 millions images. We are currently
investigating if by labeling these 80M images with a teacher, it is possible to
train shallow models with no convolutional or pooling layers to mimic deep
convolutional models.

This paper focused on training the shallowest-possible models to mimic deep
models in order to better understand the importance of model depth in learning.
As suggested in Section~\ref{sec:parallel}, there are practical applications of
this work as well: student models of small-to-medium size and depth can be
trained to mimic very large, high accuracy deep models, and ensembles of deep models, thus yielding better
accuracy with reduced runtime cost than is currently achievable
without model compression. This approach allows one to adjust flexibly the
trade-off between accuracy and computational cost.

In this paper we are able to demonstrate empirically that shallow models can,
{\em at least in principle}, learn more accurate functions without a large
increase in the number of parameters.  The algorithm we use to do
this---training the shallow model to mimic a more accurate deep model, however,
is awkward. It depends on the availability of either a large unlabeled data set
(to reduce the gap between teacher and mimic model) or a teacher model of very
high accuracy, or both. Developing algorithms to train shallow models of high
accuracy directly from the original data without going through the intermediate
teacher model would, if possible, be a significant contribution.  

\vspace{-6pt}

\section{Conclusions}

We demonstrate empirically that shallow neural nets can be trained to achieve
performances previously achievable only by deep models on the TIMIT phoneme
recognition and CIFAR-10 image recognition tasks.  Single-layer fully-connected
feedforward nets trained to mimic deep models can perform similarly to
well-engineered complex deep convolutional architectures. The results suggest
that the strength of deep learning may arise in part from a good
match between deep architectures and current training procedures, and that it may be possible
to devise better learning algorithms to train more accurate shallow
feed-forward nets. For a given number of parameters, depth may make learning
easier, but may not always be essential.

\vspace{10pt}

{\bf Acknowledgements} \hspace{2pt}We thank Li Deng for generous help with TIMIT, Li Deng 
and Ossama Abdel-Hamid for code for the TIMIT convolutional model, Chris Burges, Li Deng, Ran Gilad-Bachrach, Tapas Kanungo
and John Platt for discussion that significantly improved this work, and Mike 
Aultman for help with the GPU cluster.

\bibliography{compression}
\bibliographystyle{plainnat}

\end{document}